\documentclass[dvipsnames,format=sigconf]{acmart}
\copyrightyear{2024}
\acmYear{2024}
\setcopyright{acmlicensed}\acmConference[GECCO '24]{Genetic and Evolutionary Computation Conference}{July 14--18, 2024}{Melbourne, VIC, Australia}
\acmBooktitle{Genetic and Evolutionary Computation Conference (GECCO '24), July 14--18, 2024, Melbourne, VIC, Australia}
\acmDOI{10.1145/3638529.3654135}
\acmISBN{979-8-4007-0494-9/24/07}

\usepackage{multirow}
\usepackage{tablefootnote}
\usepackage{algorithm2e}
\RestyleAlgo{ruled}
\AtBeginDocument{%
  \providecommand\BibTeX{{%
    \normalfont B\kern-0.5em{\scshape i\kern-0.25em b}\kern-0.8em\TeX}}}

\begin{document}


\title{Effective Adaptive Mutation Rates for Program Synthesis}

\author{Andrew Ni}
\email{ani24@amherst.edu}
\orcid{0000-0003-3480-4536}
\affiliation{%
  \institution{Amherst College}
  \city{Amherst}
  \state{MA}
  \country{USA}
  \postcode{01002}
}

\author{Lee Spector}
\email{lspector@amherst.edu}
\affiliation{%
  \institution{Amherst College}
  \city{Amherst}
  \state{MA}
  \country{USA}
  \postcode{01002}
}

\renewcommand{\shortauthors}{Ni et al.}

\begin{abstract}
  The problem-solving performance of many evolutionary algorithms, including genetic programming systems used for program synthesis, depends on the values of hyperparameters including mutation rates. The mutation method used to produce some of the best results to date on software synthesis benchmark problems, 
  Uniform Mutation by Addition and Deletion (UMAD), adds new genes into a genome at a predetermined rate and then deletes genes at a rate that balances the addition rate, producing no size change on average. While UMAD with a predetermined addition rate outperforms many other mutation and crossover schemes, we do not expect a single rate to be optimal across all problems or all generations within one run of an evolutionary system. However, many current adaptive mutation schemes such as self-adaptive mutation rates suffer from pathologies like the vanishing mutation rate problem, in which the mutation rate quickly decays to zero. We propose an adaptive bandit-based scheme that addresses this problem and essentially removes the need to specify a mutation rate. Although the proposed scheme itself introduces hyperparameters, we either set these to good values or ensemble them in a reasonable range. Results on software synthesis and symbolic regression problems validate the effectiveness of our approach.
\end{abstract}





\maketitle

\section{Introduction}
Genetic Programming (GP) is a problem-solving tool that uses concepts from biological evolution to find a program that solves a specific target problem. When applied to software synthesis, GP evolves a population of programs to satisfy a set of user-defined test cases, and successful individuals are then tested for generalization to a set of previously unseen test cases. 


Many of the best results to date on benchmark program synthesis problems have been produced by PushGP, a stack-based GP system that uses linear genomes \cite{Spector2002,helmuth2018program, boldi2023informed, helmuth2015general, helmuth2021psb2}.
While many mutation and crossover schemes have been tested for program synthesis with PushGP, the variation scheme that has produced the best results recently is Uniform Mutation by Addition and Deletion (UMAD) \cite{helmuth2018program}. In contrast to uniform mutation, which replaces genes with random genes with some probability, UMAD decouples the replacement step into an addition step and a deletion step. Genes are first added into a  linear genome at a certain rate, and then deleted from the augmented genome at a rate that balances out the addition. This makes mutation  more flexible compared to random replacement, and results in better problem solving performance. While the authors show that reasonable performance can be obtained with a relatively wide range of mutation rates, and that a UMAD rate of 0.1 generally works well for many problems, they find that doubling or halving the mutation rate can sometimes improve or greatly hinder the performance of the GP system. 

Since the aim of software synthesis is to find a single solution to a set of test cases, in practice we expect end users to conduct a single, or at most a few PushGP runs until a single solution to their problem has been obtained. At that point the user will have no interest in conducting many more runs on the same problem in order to tune hyperparameters. Therefore, in order to tune hyperparameters to each individual problem, we must use an adaptive method to find optimal hyperparameter values as a run proceeds. 

In the process of biological evolution, the mutation rate of organisms' DNA is evolved alongside the organisms themselves \cite{metzgar2000evidence}. Self-Adaptation of Mutation Rates (SAMR) is an adaptive technique inspired by the biological evolution of mutation rates. In its first formulation by B\"ack \cite{back1992self}, mutation rates were represented alongside the genome itself as a bitvector. During variation, the mutation rate section of the bitvector was decoded, and then the decoded rate was used to mutate the entire genome, including the mutation bitvector. While this scheme is directly analogous to the mutation of DNA and its repair mechanisms in biological evolution, recent research in real-coded systems have shifted to using real values to represent mutation rates, and using a special meta mutation rate to vary the mutation rate. 

However, in contrast to the sheer diversity produced by biological evolution, these SAMR schemes often struggle with premature convergence due to the vanishing mutation rate problem \cite{870276, clune2008natural, 10.1145/2001576.2001699}. As most mutations in a difficult problem are deleterious, the only individuals which can survive for many generations in a row are those with very small mutation rates. 

Here we propose an adaptive controller based on multi-armed bandits and tile-codings \cite{sutton2018reinforcement} to adapt the UMAD rate to each software synthesis problem. To address the vanishing mutation rate problem, we use extreme value-based credit assignment \cite{10.1145/1143997.1144205, 10.1007/978-3-540-87700-4_18}, which allows mutation operators to be mostly deleterious as long as they occasionally produce very beneficial mutations. The use of tile-codings enables us to take advantage of the continuous domain of mutation rates, generalizing the credit assignment to a range of mutation rates for improved sample complexity. While our work is mainly focused on the current best mutation operator for software synthesis in PushGP, this scheme is not tied to a specific genetic operator and can be used to adapt other hyperparameters beyond the UMAD rate.

This paper is organized as follows: In section 2, we give a brief overview of the PushGP system for software synthesis and the UMAD mutation operator, as well as various adaptive mutation rate schemes in the literature. In section 3, we present our adaptive mutation rate scheme. In section 4, we present our experiments building up from initial validation experiments to the final performance on software synthesis and symbolic regression problems.

\section{Background and Related Work}

In this section, we give a brief overview of the PushGP system and the UMAD mutation operator, as well as a brief review on the strategies used for adaptive mutation rates.

\subsection{PushGP}
PushGP\cite{Spector2002, spector2004push} is a linear-genome GP system that evolves Push programs for solving software synthesis problems. 
Push \cite{spector2004push} is a stack-based language with a separate stack for each datatype, including one stack for the program code itself. Instructions in the program code will read off one or more values from the top of one or more datatype stacks, and push one or more values onto one or more datatype stacks. When there are not enough items on the required datatype stacks, instructions can no-op, producing no change to the program state. At the end of execution, the output of the program is simply the top one or more items from a predetermined datatype stack (e.g. the top integer for the {\tt Vector Average}), or a special token if the requested stack does not have enough elements. With the addition of instructions for moving items from the middle of a stack to the top of the stack, as well as code-modifying commands to allow for complex control structures, the Push language becomes Turing-complete. While Push programs have a hierarchical structure, they are constructed from linear plushy genomes, where genes encode the instructions and constants in a Push program, as well as the opening and closing of nested Push code blocks. 

\subsection{UMAD}

Uniform Mutation by Addition and Deletion (UMAD) \cite{helmuth2018program} is a mutation operator for variable-length, symbolic, linear genomes such as those evolved using the PushGP system. The key concept of UMAD is to extend the expressibility of traditional {\it replacement} based mutation operators by separating the deletion of an existing gene from the addition of a new gene. In UMAD, more specifically size-neutral UMAD, new genes are first inserted into a genome at a predetermined rate, usually set to 0.1. Then, genes are deleted from this augmented genome at a rate that brings the expected length of the mutated genome back to the length of the starting genome. For an addition rate of $\mu$, the deletion rate needed to balance the addition rate is $\frac{\mu}{1+\mu}$.  In general, when we refer to the UMAD rate, we mean the addition rate, and leave the deletion rate to be implicitly determined. In this paper, we extend UMAD to be defined for all positive-valued rates as follows: given a UMAD rate $\mu$, during the addition step, for every instruction in the current genome we add $\lfloor\mu\rfloor$ new instructions with probability $\{\mu\} = \mu-\lfloor\mu\rfloor$, and $\lfloor\mu\rfloor+1$ new instructions with probability $1-\{\mu\}$. The deletion probability and deletion step remain the same as before. UMAD with a rate of 0.1 has been shown to outperform or equal a variety of other mutation operators and combinations of mutation and crossover operators on software synthesis problems using the PushGP system. However, even though Helmuth et al. \cite{helmuth2018program} show that the UMAD rate is robust and performs well at different rate values, there is no single mutation rate that works optimally across all problems, and doubling or halving the default mutation rate can sometimes improve performance. We hypothesize that more ``fundamentals-based" problems, in which a clever solution idea is the crux of the problem, may require a larger UMAD rate to quickly search over the possible solution ideas. In contrast, more ``implementation-based" problems, which have many tricky details and edge cases that need to be carefully considered, may require a lower UMAD rate for a slow, methodical descent to a solution.

\subsection{Adaptive Mutation Rates}
Research on mutation rates is an extensively studied subfield of genetic algorithms \cite{10.1145/2996355, kumar2022effective, marsili2000adaptive, Maschek2010, 10.1007/3-540-61286-6_141, back1992self}. As in Aleti et al.\cite{10.1145/2996355}, we can classify mutation rate schemes into the {\it fixed}, {\it self-adaptive}, and {\it adaptive} categories. 

In the fixed mutation rate scheme the mutation rate remains constant over all problems and all generations. While hyperparameter optimization can be used to find the optimal fixed mutation rate, research has also shown that the optimal mutation rate can change over the course of evolution \cite{10.1145/2001858.2001945, 6788145}, making any fixed mutation rate suboptimal. The current mutation scheme used in PushGP is that of a fixed UMAD mutation rate, with a default rate of 0.1.

The self-adaptive mutation rate (SAMR) scheme aims to simultaneously evolve both individuals and their mutation rates \cite{meyer2007self}. In SAMR, each individual is associated with a mutation rate. During selection, individuals together with their mutation rates are chosen based on the individual's fitness function. During variation, each individual's genome is varied according to their mutation rate, and their mutation rate according to some preset meta-mutation rate. The SAMR scheme is attractive because of its elegance and its similarity to biological evolution \cite{10.1145/3205455.3205569}. However, SAMR schemes often suffer from the vanishing mutation rate problem \cite{870276, clune2008natural, 10.1145/2001576.2001699}, in which mutation rates decay to 0 and cause premature convergence. This is often attributed to the observation that SAMR schemes are myopic, only optimizing in the short term \cite{kumar2022effective}. The extent to which this short-term optimization is beneficial or detrimental can be dependent on the problem domain, the selection strength, or the solution representation \cite{870276}. However, in general SAMR struggles with more difficult problems that require optimizing for long-term improvement, often producing mutation rates which are far below optimal \cite{clune2008natural}. Some strategies for alleviating this pathology include limiting the mutation rate to within a certain interval \cite{10.1007/3-540-61286-6_141}, adding a constant to the mutation rate during variation \cite{10.1145/2001576.2001699}, or high selection pressure \cite{Maschek2010}. 

The adaptive mutation rate scheme attempts to explicitly optimize the mutation rate for better performance. Various adaptive mutation rate algorithms have been proposed to counteract the vanishing mutation rate problem. Fialho et al. \cite{10.1007/978-3-540-87700-4_18} propose to select mutation rates based on their extreme value statistics, using a dynamic multi-armed bandit and the upper confidence bound algorithm \cite{Auer2002} to balance exploration and exploitation. In contrast to the short-sightedness of SAMR, they directly optimize for the long-term effectiveness of mutation rates by choosing mutation rates with the best maximum improvement over many sampled individuals. On the other hand, Kumar et al. \cite{kumar2022effective} propose to use a coevolutionary scheme to optimize the mutation rate, which enables them to exploit the continuous nature of mutation rates. At each generation, they assign each mutation rate to a group of selected parents, producing a set of children. From that set of parent-child pairs, they then calculate the fitness of the mutation rate as the best change in fitness from parent to child in the group. They show that this maximum value-based credit assignment is effective at avoiding premature convergence compared to SAMR, and is able to converge precisely to the optimal mutation rate. 

Our work lies in the category of {\it adaptive} mutation rate schemes, and is adjacent to both the aforementioned adaptive schemes. Like Fialho et al. \cite{10.1007/978-3-540-87700-4_18}, we use a windowing method to track the maximum change in fitness obtained by a mutation rate over some number of sampled individuals, and also use a multi-armed bandit controller to balance exploration and exploitation. However, in order to take advantage of the continuous domain, we propose to use tile codings \cite{sutton2018reinforcement} to learn the bandit controller's weights. Similar to  Kumar et al.'s \cite{kumar2022effective} use of a meta-mutation rate to explore the range of possible mutation rates, we use sampling noise to choose mutation rates located around the current best rate. However, their meta-evolutionary scheme is tied to the underlying evolutionary scheme, as the suggested population size of the mutation rates approaches $N^{\frac{3}{4}}$ at large $N$ for an underlying population size $N$. This ties the sampling horizon of their max-improvement credit assignment to the population size, and makes their scheme less suited to algorithms that sample only a few individuals at a time, such as steady-state GAs or hill climbers. In contrast, our bandit scheme is able to optimize for maximum improvement over an arbitrary number of sampled individuals, and only needs the underlying scheme to repeatedly generate and evaluate children from selected parents.



\section{Methods}

In this section, we present our adaptive scheme for controlling the UMAD rate. Our method uses multi-armed bandits and tile codings for better sample complexity, and optimizes for the expected maximum improvement in fitness over many mutations, which we find to be a better indicator of the problem-solving performance of a mutation rate.

\subsection{Reward Function}

Lexicase selection\cite{6920034} is a selection algorithm for multiobjective genetic algorithms that has been shown to outperform selection methods based on aggregated fitness measures on program synthesis problems. The success of lexicase selection is commonly attributed to its ability to select ``elites," individuals which may have poor overall performance but which have excellent performance on a subset of the training cases\cite{Helmuth2020, 10.1145/3321707.3321875}. To encapsulate this idea of focusing on the individual's best performances,  we optimize the UMAD rate with respect to the symmetrical log scale-transformed error function. For each error value $x_i$ in an individual's error vector, we compute the transformed error on that test case as 
\begin{equation}
f(x_i)=\text{sgn}(x_i)\log(c+|x_i|)
\end{equation}

We base our error transformation off of the log function so that very poor performances on certain test cases will not overwhelm good performances on other test cases, following the lexicase idea. The parameter $c$ roughly represents the resolution of our proxy error function, as it behaves linearly in the region $[-c,c]$ and logarithmically outside. Therefore, we set it equal to the lowest nonnegative error we expect to see from the system. Since the function minimization problems have one-dimensional error vectors, we do not use this transformation in those experiments. For more details on the parameter values used, see appendix D.

Given a parent with errors $[e_0,\cdots,e_m]$ that was mutated to produce a child with errors $[e_0',\cdots,e_m']$, we compute the immediate reward obtained as the amount of improvement (decrease) in the average transformed error from parent to child 

\begin{equation}\label{eq:1}
    r = \frac{1}{m}\sum_{i=0}^m\log(1+e_i)-\frac{1}{m}\sum_{i=0}^m\log(1+e_i')
\end{equation}

Similarly to Fialho et al.\cite{10.1007/978-3-540-87700-4_18}, we track the maximum reward $r_{\text{max}}$ obtained over the last $len\_history$ samples from the same UMAD rate range and use that value to update our adaptive controller.

\subsection{Multi-Armed Bandits}

Multi-armed bandits\cite{sutton2018reinforcement} are one commonly used controller for adaptive mutation rates\cite{10.1007/978-3-540-87700-4_18}. The (stationary) multi-armed bandit problem is formulated as such\cite{berry1985bandit}: Given k slot machines, each with an unknown payout distribution, and a maximum number of pulls of a slot machine, how do we allocate pulls to each slot machine to maximize our expected total payout? In our case, the slot machines correspond to different UMAD rate ranges and the payouts to the reward function defined above. Different strategies have been devised for balancing exploration and exploitation in multi-armed bandits, including simple strategies like epsilon-greedy or boltzmann exploration as well as more complex strategies with better theoretical properties like upper confidence bound (UCB) based algorithms\cite{kuleshov2014algorithms}. In this work, we use an epsilon-greedy strategy  combined with sampling noise for exploration. We anneal the epsilon value from 1 to 0.01 over the first 5 generations, at which point it is kept constant at 0.01 for the rest of the run. Our sampling noise takes the form of gaussian perturbation, where we sample UMAD rate intervals in the vicinity of the best rate interval according to a normal distribution.

\subsection{Tile Codings}

Current works using multi-armed bandits for adaptive mutation rate control do not take advantage of the continuous nature of this domain, instead partitioning possible rates into separate intervals\cite{10.1007/978-3-540-87700-4_18, 10.1145/1143997.1144205}. Intervals which are too large will mask the performance of good mutation rates with the poor performance of significantly different mutation rates. On the other hand, intervals which are too small suffer from poor sample complexity, as each interval needs to be sampled many times in order to obtain an accurate estimate of the expected reward. In order to combine generalization with precision, we use tile-codings\cite{sutton2018reinforcement} to learn the expected $r_{\text{max}}$ associated with different UMAD rates. In tile codings, we create many tilings of our desired parameter range with different tile offsets and widths. When we observe a reward associated with a specific point, we update all the tiles covering that point using that reward. Then, when we want to assess the value of a point, we average the values of all the tiles covering that point. With enough random tilings, we can distinguish between very small UMAD rate ranges with high precision, while each sampled reward will still update tiles covering a large range of UMAD rates for good generalization and low sample complexity.

Given the very stochastic rewards in our system, we use SGD with nesterov momentum\cite{pmlr-v28-sutskever13} as well as ensembling to stabilize learning. A tile coding defined on the range $[l,r]$ with tile offset $o$ and width $w$ will track the value (expected $r_{\text{max}}$) and momentum associated with the intervals $[l, l+o)$, $[l+o, l+o+w)$, $\cdots$, $[l+o+\lfloor\frac{r-l-o}{w}\rfloor\cdot w, r)$. When a tile coding with values $[v_0,\cdots,v_n]$ and momentum $[m_0,\cdots,m_n]$ observes a reward $r$ at position $x$, it will calculate the index of the tile that covers $x$ as $i=\lfloor\frac{x-o-l}{w}\rfloor+1$. Then, using the learning rate $\gamma$ and momentum factor $\mu$, it will calculate the gradient $g = 2(v_i-r)$, update the momentum $m_i\leftarrow \mu m_i+g$, and update the value using the updated momentum $v_i\leftarrow v_i-\gamma(g+\mu m_i)$. 

In practice, we create multiple multi-armed bandits each with a randomized learning rate. During variation, a multi-armed bandit is chosen at random to sample for a mutation rate. Then, once we have computed the immediate reward $r$ for that sampled mutation rate, all of the bandit controllers are updated with that information. 

\subsection{Adaptive UMAD Rate}

We combine the previous sections into an adaptive controller for UMAD rates in problem synthesis. The algorithm for sampling a UMAD rate from a single bandit is given in Algorithm 1, and the algorithm for updating a single bandit is given in Algorithm 2. In practice, we have multiple bandits, updating them all with the same rewards and randomly choosing a bandit to sample from.

\begin{algorithm}
\caption{Sampling UMAD rate from a single bandit}\label{alg:one}
\KwData{\begin{itemize}
    \item [$\bullet$] Search range $[l, r]$
    \item [$\bullet$] Search resolution $res$
    \item [$\bullet$] Exploration noise $\sigma$
    \item [$\bullet$] Epsilon-greedy exploration  rate $\epsilon$
    \item [$\bullet$] Number of tile codings $num\_codings$
    \item[$\bullet$] Base coding $base\_coding = \text{Coding}(l,r,0,res)$
    \item[$\bullet$] Tile codings $tile\_codings = \{\text{Coding}_i(l, r, o_i, w_i)\}$
    \item [$\bullet$] $V=\{v_{i,j}\}$ the value of the $j^{th}$ tile in the $i^{th}$ tile coding
\end{itemize}}
\KwResult{$\rho$, the UMAD rate to use for mutation}
$weights$ $\longleftarrow$ []

$num\_base\_tiles\longleftarrow \lfloor\frac{r-l}{res}\rfloor$

\tcc{Compute tile weights in the base coding}
\For{$i\leftarrow 0$ \KwTo $num\_base\_tiles$}{
    \tcc{Average the associated values in each tile coding}
    $total\longleftarrow0$
    
    \For{$j\leftarrow0$ \KwTo $num\_codings$}{
        $idx\longleftarrow\lfloor\frac{i\cdot res-o_j}{w_j}\rfloor+1$
        
        $total+=v_{j,idx}$
    }
    $weights$.{\tt append}($\frac{total}{num\_codings}$)
}

\eIf{(rand) < $\epsilon$}{
\tcc{Sample a random tile}
$best\_tile\sim \lfloor\mathcal{U}([0,num\_base\_tiles])\rfloor$
}{
\tcc{Sample around the best tile}
$best\_tile\sim$ {\tt argmax}($weights$) $+\lfloor\mathcal{N}(0,\sigma)\rfloor$

$best\_tile\longleftarrow {\tt max}({\tt min}( best\_tile, num\_base\_tiles-1), 0)$

}

\tcc{Uniformly sample a log UMAD rate from within the selected tile}

$log\_rate\sim \mathcal{U}([l+res\cdot best\_tile,l+res\cdot (best\_tile+1)])$

\tcc{Convert from log UMAD rate to UMAD rate}

\Return{$e^{log\_rate}$}

\end{algorithm}

\begin{algorithm}
\caption{Updating a single bandit}\label{alg:two}
\KwData{\begin{itemize}
    \item [$\bullet$] Search range $[l, r]$
    \item [$\bullet$] Search resolution $res$
    \item [$\bullet$] Learning rate $\gamma$
    \item [$\bullet$] Momentum factor $\mu$
    \item [$\bullet$] $len\_history$ the number of past rewards to max over
    \item[$\bullet$] $tile\_coding = \text{Coding}(l, r, o, w)$ the tile coding
    \item [$\bullet$] $V=\{v_{i}\}$ the value of the $i^{th}$ tile in the tile coding
    \item [$\bullet$] $M=\{m_{i}\}$ the momentum of the $i^{th}$ tile in the tile coding
    \item [$\bullet$] $R=\{r_{i}\}$ the history of rewards obtained by tiles in the tile coding, represented as deques with max length $len\_history$
    \item [$\bullet$] $x$ the sampled UMAD rate
    \item [$\bullet$] $E=\{e_i\}$ the parent error vector
    \item [$\bullet$] $E'=\{e_i'\}$ the child error vector
\end{itemize}}
\KwResult{\begin{itemize}
\item [$\bullet$] \{$v_i$\} the updated tile coding values
\item [$\bullet$] \{$m_i$\} the updated tile coding momentum
\end{itemize}}
\tcc{Locate the tile containing the UMAD rate}
$idx \longleftarrow\lfloor\frac{\log{x}-l-o}{w}\rfloor+1$

\tcc{Compute immediate reward}
$reward = (\log(1+E')-\log(1+E)).{\tt mean}()$

\tcc{Compute max over reward history}
$r_{idx}.{\tt push}(reward)$

$max\_reward\longleftarrow\max(r_{idx}[0],\cdots,r_{idx}[len\_history-1])$

\tcc{Compute gradient and update parameters}
$g\longleftarrow 2(v_{idx}-max\_reward)$

$m_{idx}\longleftarrow \mu\cdot m_{idx}+g$

$v_{idx}\longleftarrow v_{idx}-\gamma(g+\mu\cdot m_{idx})$

\Return{$\{m_i\}$, $\{v_i\}$}

\end{algorithm}

\section{Experiments}

In this section we present our experiments validating the effectiveness our algorithm, exploring the fitness landscape of software synthesis problems, and demonstrating the performance of our controller on genetic programming and symbolic regression problems. 

\subsection{Function Minimization}

We first validate the effectiveness of our algorithm on some function minimization problems. In these problems, a real-valued n-dimensional vector is evolved to minimize a synthetic fitness function. We use the common test functions {\tt Ackley}, {\tt Greiwank}, \\{\tt Rastrigin}, {\tt Rosenbrock}, {\tt Sphere}, and {\tt Linear} \cite{kumar2022effective, 4983112, simlib}. However, due to differences in genetic algorithm hyperparameters such as the truncation size, our results are not comparable to those in the literature. The test problem definitions can be found in appendix A. We run each problem with a 100-dimensional test function using a population size of 100 + 1 elite, and a generation limit of 1000. We initialize the GA population according to the normal distribution $\mathcal{N}({\bf 0},\sigma^2{\bf I})$ where the standard deviation $\sigma$ is on roughly the same order of magnitude as the dimensions recommended in \cite{simlib}, or 1 if such a recommendation is not available. For specific details, see appendix A. The one exception is the {\tt Linear} problem which is only run for 100 generations. We compare our adaptive bandit-based controller against the GESMR, SAMR, and LAMR-100 mutation rate schemes from Kumar et al.\cite{kumar2022effective}. GESMR is the the novel evolutionary mutation rate adaptation proposed in Kumar et al.\cite{kumar2022effective}, which coevolves a population of mutation rates alongside the main population. SAMR is a simple self-adaptive mutation rate scheme that attaches mutation rates to individuals and evolves mutation rates and genomes together. The LAMR-100 scheme, which is our ``oracle," determines the optimal mutation rate every 100 generations by ``looking ahead," running each mutation value in a preset range for 100 generations and choosing the best one. Therefore LAMR-100 is able to directly optimize for future behavior. For these experiments, LAMR-100 searches over a log-spaced range of values from $10^{-3}$ to $10^0$. To showcase the sample efficiency of our controller, we let the bandit controller search over a very large log-range of mutation rates from $-100$ to $100$. In addition, we use an amount of sampling noise roughly equal to the meta mutation strength of GESMR. For specific parameter details, see appendix D. All results are averaged over 50 runs. 

\begin{figure}
    \centering
    \includegraphics[width=0.45\textwidth]{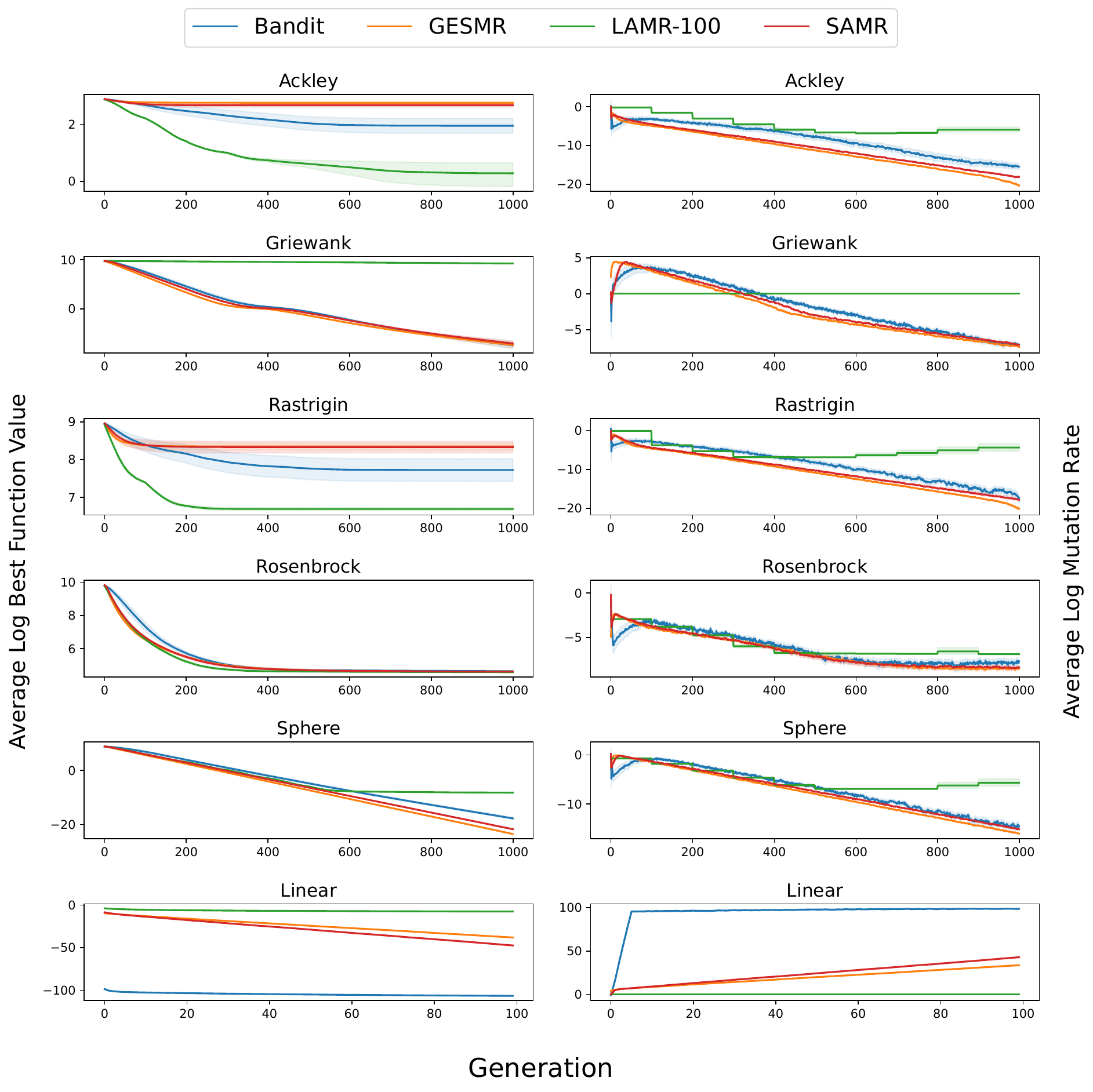}
    \caption{Performance of four adaptive mutation rate schemes on function minimization problems. 95\% confidence intervals obtained by bootstrapping. The bandit controller is competitive with GESMR, with slightly slower adaptation on easy problems like {\tt Rosenbrock} and {\tt Sphere}, but better final fitness on rugged domains like {\tt Ackley} and {\tt Rastrigin}.}
    
    \label{fig:funcmin}
\end{figure}

\begin{table}
    \centering
    \caption{Average final function value of adaptive mutation rate schemes on function minimization problems. The LAMR-100 oracle is not used for comparison. The best non-oracle value is shown in bold. Results significantly worse than the best scheme with $p<0.05$ using Welch's t-test  are underlined.}
    \begin{tabular}{c|c|c|c|c}\toprule
        Problem & Bandit & GESMR & SAMR & LAMR-100\\\midrule
        {Ackley} & {\bf 10.1} & \underline{15.8} & \underline{14.8}  & 2.0\\
        {Griewank} & {\bf 4.69e-3} & 4.95e-3 & 5.24e-3 & 1.02e+4 \\
        {Rastrigin} & {\bf 3686} & 4505 & \underline{4772} & 815\\
        {Rosenbrock} & 105 & {\bf 102} & {\bf 102} &98\\
        {Sphere} & \underline{5.56e-8} & {\bf 6.86e-11} & \underline{4.52e-10} & 2.74e-4\\
        {Linear} & \bf{-2.91e+46} & \underline{-1.12e+17} & \underline{-2.10e+21} & -1949\\\bottomrule
    \end{tabular}
    
    \label{tab:funcmin_tab}
\end{table}

The results of our experiment is displayed in figure \ref{fig:funcmin}. Despite the wide range of mutation standard deviations searched by our controller, we still have high precision and good sample complexity due to our use of tile codings, which efficiently exploit the continuous nature of mutation rates. 
As we have set $len\_history$ to 100 in this paper, our bandit controller optimizes for more long-term performance than GESMR. This is because, following the trend reported in Kumar et al.\cite{kumar2022effective}, we have set the GESMR meta population size to 10. Therefore, the fitness function of a mutation rate is the maximum improvment in fitness of 10 sampled individuals for GESMR, whereas we choose mutations based on the expected maximum over 100 individuals. For this reason, our controller outperforms GESMR which outperforms SAMR on more rugged domains due to their better ability to ignore local minima in favor of global minima. However, since our bandit controller directly estimates this order statistic empirically, it requires more samples to learn and takes more time to adapt the mutation rate. This results in weaker performance on easier problems like {\tt Rosenbrock} and {\tt Sphere}. 

Since the LAMR-100 oracle optimizes for performance over 100 generations, or 10,000 individuals, it is able to achieve the best performance on the rugged problems {\tt Ackley} and {\tt Rastrigin}. However, the limitations of the LAMR-100 method are also clear. Because LAMR-100 only searches over mutation rates from $10^{-3}$ to $10^0$, it is unable to perform as well as the other methods on the Linear and Griewank problems, which both require higher mutation rates. In addition, the sampled mutation rates flatten out at later generations on the {\tt Ackley}, {\tt Rastrigin}, and {\tt Sphere} problems at the minimum value available to LAMR-100, $10^{-3}$, whereas the optimal mutation rate likely continues decreasing according to the trend seen in earlier generations. 

Out of the six problems tested, the {\tt Linear} problem is unique in its unbounded and smooth fitness function. While GESMR and SAMR will eventually outperform our bandit controller on the {\tt Linear} problem due to their unbounded growth of mutation rates, our bandit controller is able to learn much more quickly than GESMR or SAMR, essentially achieving its maximum mutation rate within the 5 exploration generations.

\subsection{Software Synthesis}

For our experiments in software synthesis, we use the {\tt propeller} implementation of PushGP\footnote{https://github.com/lspector/propeller} and tackle several problems from the PSB1 and PSB2 benchmarks. These problems were chosen to represent a range of difficulties and consist of three problems from each benchmark suite.

\subsubsection{Fitness Landscape}

Compared to the continuous test functions, software synthesis is a much less regular domain. There are fewer beneficial mutations, and the changes in total fitness from parent to child can vary significantly. We therefore expect SAMR to suffer from the vanishing mutation rate problem on this domain. Before conducting problem-solving expeirments, we first validate our choice of a max-value based credit assignment $r_{\text{max}}$ and show that it can help avoid the vanishing mutation rate problem. 

In these experiments, we run PushGP on each software synthesis problem with the default UMAD rate of 0.1. We use the default settings\cite{Helmuth2020,helmuth2021psb2,boldi2023informed}, with a population size of 1000 and generation limit of 300, and lexicase selection\cite{6920034} as the selection operator. At each generation, we additionally sample 1000 individuals using each UMAD rate $\mu\in\{0.01, 0.03, 0.1, 0.3, 1\}$. These additional individuals have no impact on the genetic algorithm, and are solely used to evaluate the performance of each mutation rate under our defined reward function. For each individual sampled this way, we compute its immediate reward as described in equation \ref{eq:1}. For each UMAD rate, we combine all of these sampled rewards over the entire GP run into a linear array, apply 1-dimensional max pooling with kernel size $len\_history$ and stride $1$ to transform it into the maximum reward $r_{\text{max}}$, and then take an exponentially weighted average with a learning rate of 0.01 to smooth out the plot. We also do the same process without max-pooling, essentially plotting the immediate reward $r$. Finally, we also display the best log-error in the current population at each generation to get an idea of how far evolution has progressed. For all the experiments in this paper, we use $len\_history=100$. All problems were run for 3 runs each, and at each generation the average statistic was taken over all runs which had not yet found a solution.

\begin{figure}
    \centering
    \includegraphics[width=0.45\textwidth]{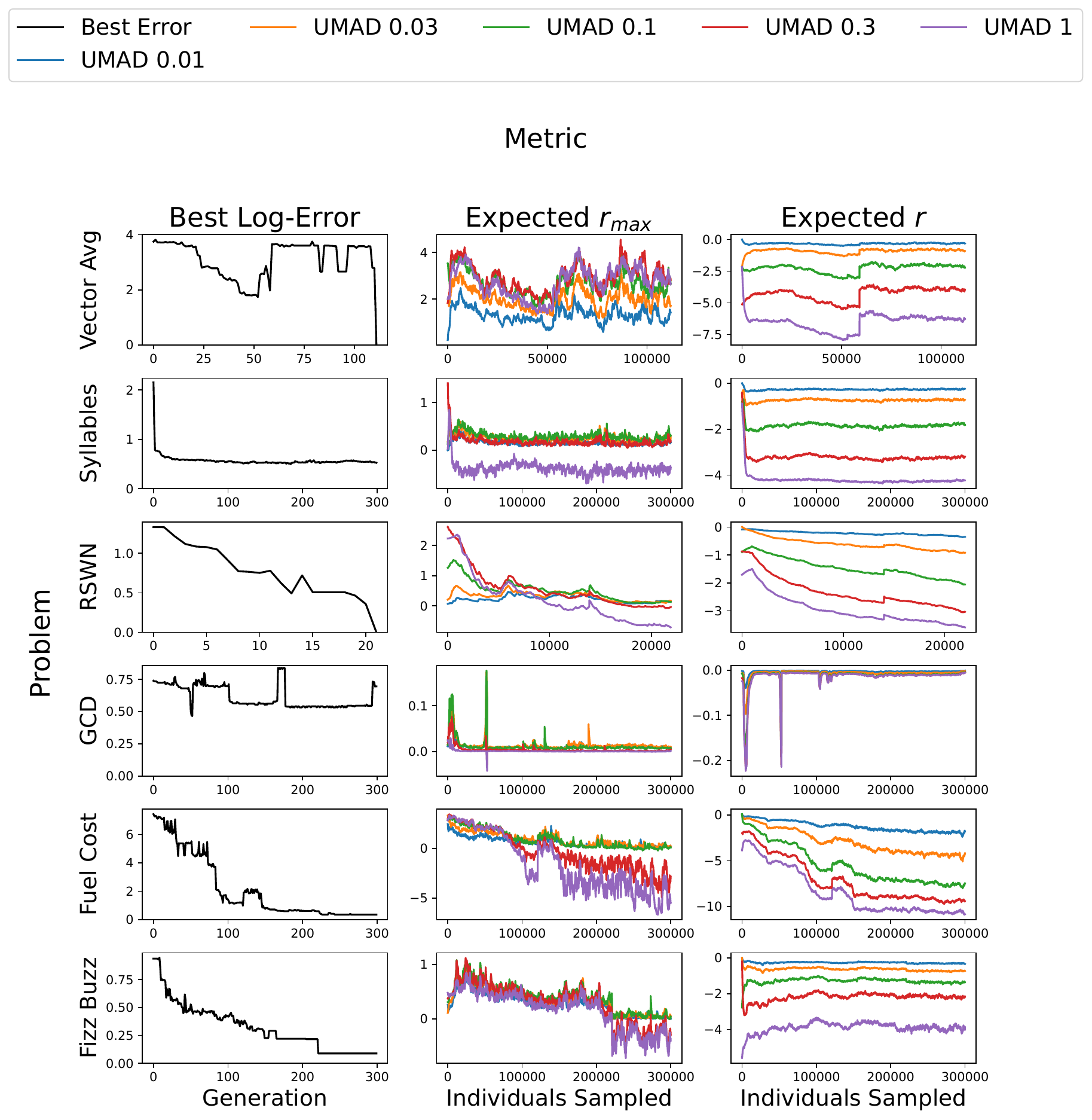}
    \caption{Rewards obtained by different UMAD rates on software synthesis problems. Naively considering only the expected reward makes the lowest UMAD rate appear to be the best. Considering the expected maximum reward over $len\_history$ samples paints a much more realistic picture.}
    
    \label{fig:landscape}
\end{figure}

The results are depicted in figure \ref{fig:landscape}. If we naively consider only the immediate reward, it would appear that on all problems, the lower the mutation rate the better. In fact, these expected rewards are all negative throughout all generations of all problems, so all the mutation rates studied have a worse expected $r$ than zero mutation. Therefore, we expect using the expected immediate reward as a metric to result in mutation rates converging to zero. However, if we instead consider the max-value based reward $r_{\text{max}}$, we see that larger mutation rates are generally better when the best solution found is still poor, but as evolution proceeds and the solutions improve, lower mutation rates become better.  We therefore expect the $r_{\text{max}}$ metric to be a much more realistic predictor of mutation rate performance, and we expect that controllers based on maximum value statistics will avoid the vanishing mutation rate problem.

\subsubsection{Software Synthesis Performance}

As in the previous section, we use the propeller implementation of PushGP and run on the same software synthesis problems with the same settings. In this section, we compare the performance of a fixed UMAD rate of 0.1 with our adaptive bandit controller as well as a self-adaptive mutation rate scheme. Due to computational limitations, instead of reporting the number of successes out of 100 as recommended in \cite{helmuth2021psb2}, we report the number of successes out of 50.

\begin{figure}
    \centering
    
    \includegraphics[width=0.45\textwidth]{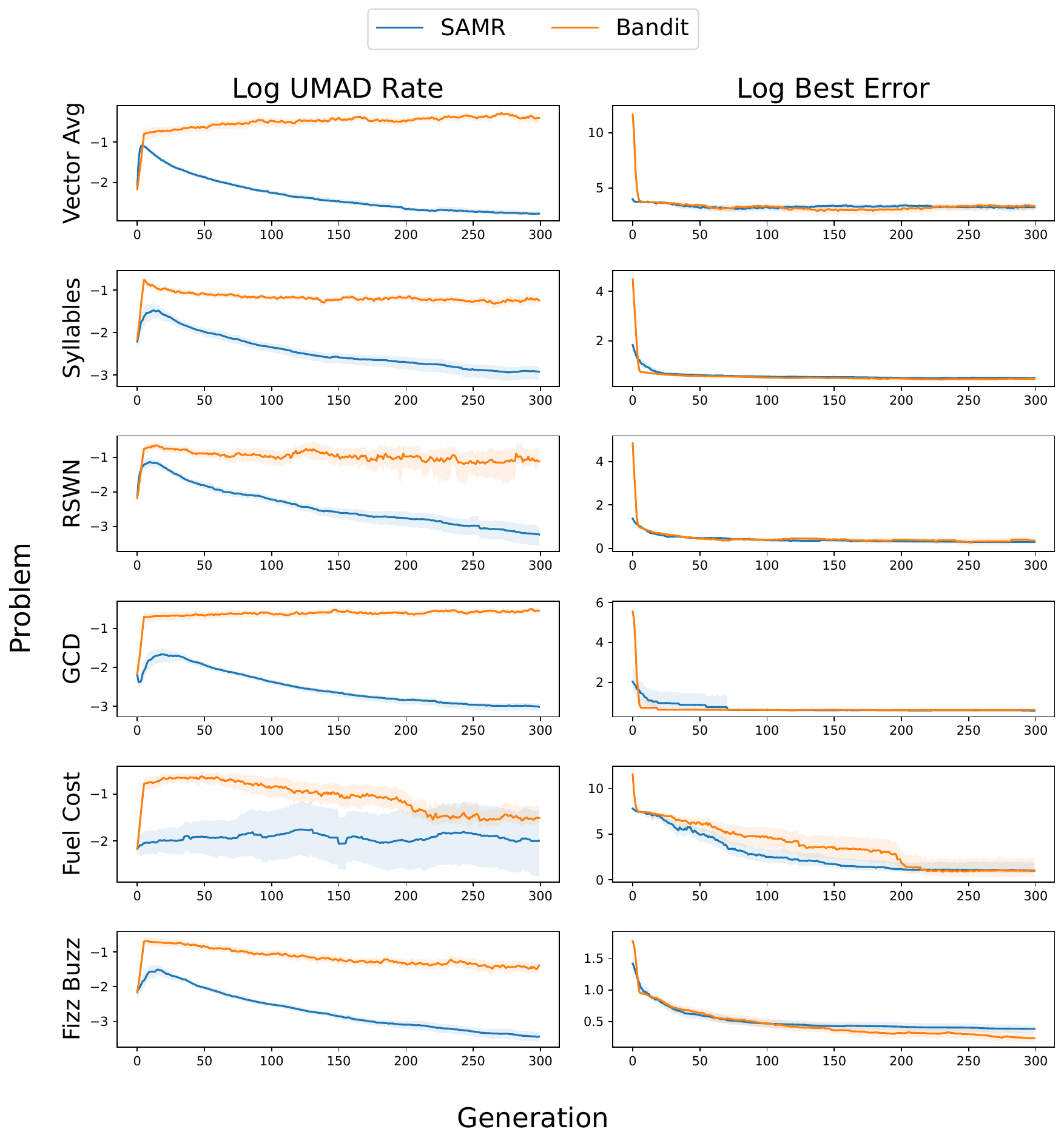}
    \caption{Best error and average log-UMAD rate sampled by three different adaptive controllers on Software Synthesis problems. 95\% confidence intervals obtained by bootstrapping. The bandit controller is able to effectively avoid the vanishing mutation rate problem.}
    
    \label{fig:gp_rates}
\end{figure}

\begin{table}
    \centering
    \caption{Number of successes out of 50 runs  of fixed and adaptive mutation rate schemes on Software Synthesis problems. The best performance among adaptive results is shown in bold. The performance of the fixed UMAD rate of 0.1 is bolded when best but does not prevent bolding of the adaptive schemes. Statistical significance was calculated using a two-proportion z-test. Results significantly worse than the bandit scheme with $p<0.05$ are underlined. No results from adaptive schemes were significantly better than the bandit scheme. }
    \begin{tabular}{c|c|c|c}\toprule
        Problem & UMAD 0.1 & Bandit & SAMR\\\midrule
        { Vector Average} & {\bf 45} & {\bf 27} & \underline{16} \\
        { Syllables} & 10 & {\bf 18} & \underline{6} \\
        { Replace Space With Newline} & {\bf 47} & {\bf 43} & \underline{23} \\
        { GCD} & {\bf 6} & 1& {\bf 2}\\
        { Fuel Cost} & 21 & {\bf 29} & 23\\
        { Fizz Buzz} & 8 & {\bf 12} & \underline{1}\\\bottomrule
    \end{tabular}
    
    \label{tab:propeller}
\end{table}

The results are shown in table \ref{tab:propeller}. Although the bandit controller is not always able to perform as well as the fixed UMAD rate, it is often able to significantly outperform the SAMR scheme. To see why, we plotted the best log-error and average log-UMAD rate of each adaptive scheme for each problem over 300 generations. As before, we average each statistic at each generation over all the runs which have not yet terminated. The results are shown in figure \ref{fig:gp_rates}. Although SAMR is able to increase the UMAD rate in the outset of evolution, the UMAD rate quickly decays to very low values, resulting in premature convergence and a low solution rate. On the other hand, our bandit controller is better able to avoid the vanishing mutation rates, chooses more optimal UMAD rates, and achieves better solution rates.

\subsection{Symbolic Regression Performance}

As in the previous section, we use the propeller implementation of PushGP. In this experiment, we run on the eight 1-dimensional synthetic regression problems from Nguyen et al.\cite{Uy2011}. The target functions are detailed in appendix B. As the details of our experiment differ from those in the literature, our results only serve to draw comparisons between the adaptive mutation rate schemes studied here. We define a successful function as one which completely replicates the given input-output pairs, to within some small constant. In addition, we use an instruction set composed of the input instruction {\tt input}, the constant {\tt 1.0}, basic arithmetic operations ({\tt +}, {\tt -}, {\tt $\times$}, {\tt $\div$}), and the additional functions ({\tt Sin}, {\tt Cos}, {\tt Log}). We protect {\tt Log} and {\tt $\div$} to return $0.0$ for undefined inputs, and clamp all numbers in the execution of our program to the interval $[-1.0\times10^6, 1.0\times10^6]$. For our test cases on problems {\tt Nguyen1} through {\tt Nguyen6}, we use a range of evenly spaced inputs from -4 to 4 with a step size of 0.1. For {\tt Nguyen7} and {\tt Nguyen8}, to avoid undefined regions of the domain, we shift the inputs to range from 0 to 8.  We find that the larger range of inputs gives more information about the shape of the function, leading to higher quality runs. We use the epsilon-lexicase selection method \cite{10.1145/2908812.2908898} and run PushGP with a population size of 1000 for 300 generations. We report our results as the number of successful runs out of 50 total.

\begin{table}
    \centering
    \caption{Number of successes out of 50 runs of fixed and adaptive mutation rate schemes on Symbolic Regression problems. UMAD 0.1 is bolded when best but does not prevent bolding of the adaptive schemes. Statistical significance was calculated using a two-proportion z-test. Results significantly worse than the bandit scheme with $p<0.05$ are underlined. No results from adaptive schemes were significantly better than the bandit scheme.}
    \begin{tabular}{c|c|c|c|c}\toprule
        Problem & UMAD 0.1 & Bandit & GESMR & SAMR \\\midrule
        Nguyen1 & {\bf 50} & {\bf 50} & 47& 47\\
        Nguyen2 & {\bf 50} & {\bf 50} & 48 & \underline{34} \\
        Nguyen3 & {\bf 48} &  {\bf 43} &\underline{12} &   \underline{9}\\
        Nguyen4 & 34 &  {\bf 43} & \underline{14} &  \underline{2} \\
        Nguyen5 & {\bf 50} & {\bf 42} & \underline{24} & \underline{23}\\
        Nguyen6 & {\bf 50} &  {\bf 30} & \underline{6} & 23\\
        Nguyen7 & {\bf 8} &  2 & 0& {\bf 3}\\
        Nguyen8 & 0 &  0 & 0 & 0\\\bottomrule
    \end{tabular}
    
    \label{tab:nguyen}
\end{table}

\begin{figure}
    \centering
    
    \includegraphics[width=0.45\textwidth]{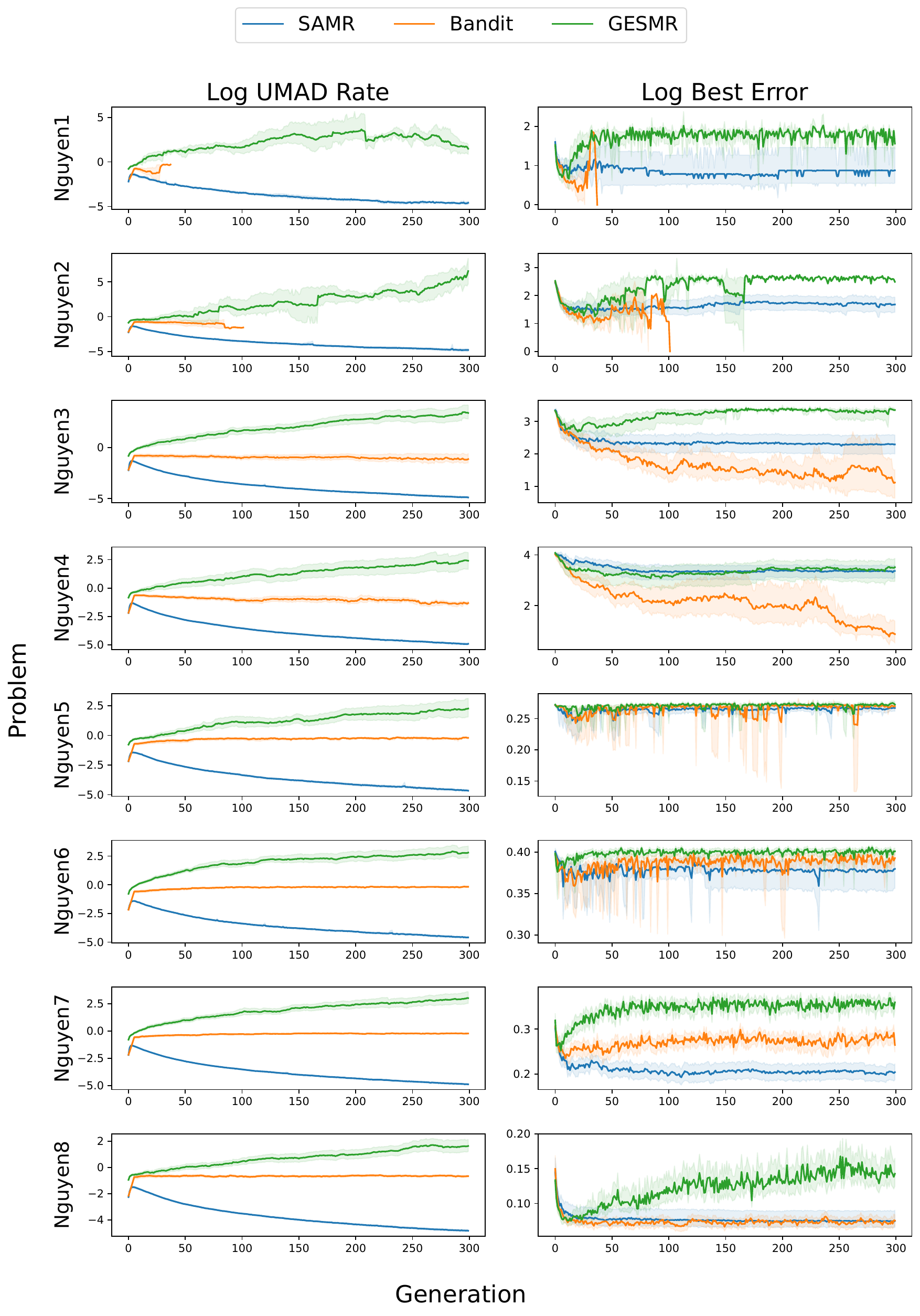}
    \caption{Best error and average log-UMAD rate sampled by three different adaptive controllers on SR problems. 95\% confidence intervals obtained by bootstrapping. SAMR suffers from a vanishing mutation rate, while GESMR suffers from an exploding mutation rate.}
    \label{fig:sr_rates}
\end{figure}

The results are displayed in table \ref{tab:nguyen}. The bandit based scheme consistently outperforms the SAMR and GESMR schemes on the symbolic regression problems, although it is not always able to perform as well as the preset UMAD rate. The best log-error and average log-UMAD rate sampled by these three adaptive methods are displayed in figure \ref{fig:sr_rates}. As before, we average each statistic at each generation over all the runs that have not yet terminated. Therefore, for problems like Nguyen1 and Nguyen2 where the bandit method always terminates within 50 or 100 generations, there is no data for later generations. In all problems, the SAMR method results in continually decreasing mutation rates which often results in premature convergence, such as in the {\tt Nguyen3} and {\tt Nguyen4} problems. On the other hand, GESMR samples ever increasing UMAD rates. We hypothesize that this is due to the low degree of differentiation between large UMAD rates. As the UMAD rate tends towards infinity, UMAD mutation does not tend towards infinitely large mutations as with gaussian mutation, but rather towards completely random mutation. Therefore, the difference between mutation rates is much less pronounced, especially for large mutation rates, in the symbolic regression domain than it is in the function minimization domain. As GESMR only utilizes local information about mutation rates, we hypothesize that it gets stuck in this uninformative region of high mutation rates and is unable to realize when significant progress has been made and large mutation rates are no longer optimal. In contrast, the bandit controller is able to escape these pathologies and robustly determine a good, moderate UMAD rate at each generation.

\section{Conclusion and Future Work}

We have presented an adaptive scheme for controlling the rate of UMAD mutation in genetic programming problems. The proposed method uses multi-armed bandits to optimize the mutation rate, tile codings to improve generalization, and optimizes for the maximum improvement over many mutations, which is a more accurate predictor of the usefulness of a given mutation rate. 

In the future, we could extend our adaptive controller to other mutation and selection operators in addition to the UMAD rate. For example, we could learn various combinations of tree-based mutation operators at various strengths for tree-based GP systems\cite{fieldguidegp}, or various bit mutation operators for systems with binary repre- sentations\cite{10.1007/978-3-540-87700-4_18}. One attractive domain for adaptive optimization is the genetic source\cite{10.1162/isal_a_00326}. An effective adaptive scheme for genetic source optimization not only has the potential for greatly improved solution rates\cite{10.1162/isal_a_00326}, but also can relieve the end user of the need to specify which datatypes and instructions are needed for each individual problem. 

Beyond genetic programming, this system could also be used to adapt hyperparameters such as learning rate or weight decay over the course of training a neural network. To do this, we frame the task of choosing hyperparameters as the maximization of the expected improvement in the training loss, which can be estimated by the improvement in training loss from the current training batch to the next. While we could also use the expected improvement of the validation loss, this formulation does not require any additional forward passes, and has almost no overhead cost. Since we are interested in the continuous improvement of a single model in this domain, we would maximize the expected reward, and not the expected maximum reward over $len\_history$ samples as we have done in the evolutionary domains.

However, the scope of possible applications of our controller is limited by the curse of dimensionality, as the number of tiles to be evaluated grows exponentially with the number of parameters to be optimized. One simple fix would be to keep a separate bandit controller for each hyperparameter. However, this assumes that the hyperparameters can be optimized independently, ignoring the possible interplay between hyperparameters. In the future, we could extend our adaptive scheme to higher dimensions, such as by replacing the tile coding with a neural network.

\begin{acks}

This work was performed in part using high-performance computing equipment at Amherst College obtained under National Science Foundation Grant No. 2117377. Any opinions, findings, and conclusions or recommendations expressed in this publication are those of the authors and do not necessarily reflect the views of the National Science Foundation. The authors would like to thank Ryan Boldi, Bill Tozier, Tom Helmuth, Edward Pantridge and other members of the PUSH lab for their insightful comments and suggestions.
    
\end{acks}

\bibliographystyle{ACM-Reference-Format}
\bibliography{bibliography}


\begin{thebibliography}{37}


\ifx \showCODEN    \undefined \def \showCODEN     #1{\unskip}     \fi
\ifx \showDOI      \undefined \def \showDOI       #1{#1}\fi
\ifx \showISBNx    \undefined \def \showISBNx     #1{\unskip}     \fi
\ifx \showISBNxiii \undefined \def \showISBNxiii  #1{\unskip}     \fi
\ifx \showISSN     \undefined \def \showISSN      #1{\unskip}     \fi
\ifx \showLCCN     \undefined \def \showLCCN      #1{\unskip}     \fi
\ifx \shownote     \undefined \def \shownote      #1{#1}          \fi
\ifx \showarticletitle \undefined \def \showarticletitle #1{#1}   \fi
\ifx \showURL      \undefined \def \showURL       {\relax}        \fi
\providecommand\bibfield[2]{#2}
\providecommand\bibinfo[2]{#2}
\providecommand\natexlab[1]{#1}
\providecommand\showeprint[2][]{arXiv:#2}

\bibitem[\protect\citeauthoryear{Aleti and Moser}{Aleti and Moser}{2016}]%
        {10.1145/2996355}
\bibfield{author}{\bibinfo{person}{Aldeida Aleti} {and} \bibinfo{person}{Irene Moser}.} \bibinfo{year}{2016}\natexlab{}.
\newblock \showarticletitle{A Systematic Literature Review of Adaptive Parameter Control Methods for Evolutionary Algorithms}.
\newblock \bibinfo{journal}{\emph{ACM Comput. Surv.}} \bibinfo{volume}{49}, \bibinfo{number}{3}, Article \bibinfo{articleno}{56} (\bibinfo{date}{Oct} \bibinfo{year}{2016}), \bibinfo{numpages}{35}~pages.
\newblock
\showISSN{0360-0300}
\urldef\tempurl%
\url{https://doi.org/10.1145/2996355}
\showDOI{\tempurl}


\bibitem[\protect\citeauthoryear{Auer, Cesa-Bianchi, and Fischer}{Auer et~al\mbox{.}}{2002}]%
        {Auer2002}
\bibfield{author}{\bibinfo{person}{Peter Auer}, \bibinfo{person}{Nicol{\`o} Cesa-Bianchi}, {and} \bibinfo{person}{Paul Fischer}.} \bibinfo{year}{2002}\natexlab{}.
\newblock \showarticletitle{Finite-time Analysis of the Multiarmed Bandit Problem}.
\newblock \bibinfo{journal}{\emph{Machine Learning}} \bibinfo{volume}{47}, \bibinfo{number}{2} (\bibinfo{date}{01 May} \bibinfo{year}{2002}), \bibinfo{pages}{235--256}.
\newblock
\showISSN{1573-0565}
\urldef\tempurl%
\url{https://doi.org/10.1023/A:1013689704352}
\showDOI{\tempurl}


\bibitem[\protect\citeauthoryear{B{\"a}ck et~al\mbox{.}}{B{\"a}ck et~al\mbox{.}}{1992}]%
        {back1992self}
\bibfield{author}{\bibinfo{person}{Thomas B{\"a}ck} {et~al\mbox{.}}} \bibinfo{year}{1992}\natexlab{}.
\newblock \showarticletitle{Self-adaptation in genetic algorithms}. In \bibinfo{booktitle}{\emph{Proceedings of the first european conference on artificial life}}. MIT press Cambridge, \bibinfo{pages}{263--271}.
\newblock


\bibitem[\protect\citeauthoryear{B{\"a}ck and Sch{\"u}tz}{B{\"a}ck and Sch{\"u}tz}{1996}]%
        {10.1007/3-540-61286-6_141}
\bibfield{author}{\bibinfo{person}{Thomas B{\"a}ck} {and} \bibinfo{person}{Martin Sch{\"u}tz}.} \bibinfo{year}{1996}\natexlab{}.
\newblock \showarticletitle{Intelligent mutation rate control in canonical genetic algorithms}. In \bibinfo{booktitle}{\emph{Foundations of Intelligent Systems}}, \bibfield{editor}{\bibinfo{person}{Zbigniew~W. Ra{\'{s}}} {and} \bibinfo{person}{Maciek Michalewicz}} (Eds.). \bibinfo{publisher}{Springer Berlin Heidelberg}, \bibinfo{address}{Berlin, Heidelberg}, \bibinfo{pages}{158--167}.
\newblock
\showISBNx{978-3-540-68440-4}


\bibitem[\protect\citeauthoryear{Berry and Fristedt}{Berry and Fristedt}{1995}]%
        {berry1985bandit}
\bibfield{author}{\bibinfo{person}{Donald~A Berry} {and} \bibinfo{person}{Bert Fristedt}.} \bibinfo{year}{1995}\natexlab{}.
\newblock \bibinfo{booktitle}{\emph{Bandit problems: sequential allocation of experiments (Monographs on statistics and applied probability)}}.
\newblock \bibinfo{publisher}{Springer}.
\newblock


\bibitem[\protect\citeauthoryear{Boldi, Briesch, Sobania, Lalejini, Helmuth, Rothlauf, Ofria, and Spector}{Boldi et~al\mbox{.}}{2024}]%
        {boldi2023informed}
\bibfield{author}{\bibinfo{person}{Ryan Boldi}, \bibinfo{person}{Martin Briesch}, \bibinfo{person}{Dominik Sobania}, \bibinfo{person}{Alexander Lalejini}, \bibinfo{person}{Thomas Helmuth}, \bibinfo{person}{Franz Rothlauf}, \bibinfo{person}{Charles Ofria}, {and} \bibinfo{person}{Lee Spector}.} \bibinfo{year}{2024}\natexlab{}.
\newblock \showarticletitle{{Informed Down-Sampled Lexicase Selection: Identifying Productive Training Cases for Efficient Problem Solving}}.
\newblock \bibinfo{journal}{\emph{Evolutionary Computation}} (\bibinfo{date}{03} \bibinfo{year}{2024}), \bibinfo{pages}{1--31}.
\newblock
\showISSN{1063-6560}
\urldef\tempurl%
\url{https://doi.org/10.1162/evco\_a\_00346}
\showDOI{\tempurl}


\bibitem[\protect\citeauthoryear{Clune, Misevic, Ofria, Lenski, Elena, and Sanju{\'a}n}{Clune et~al\mbox{.}}{2008}]%
        {clune2008natural}
\bibfield{author}{\bibinfo{person}{Jeff Clune}, \bibinfo{person}{Dusan Misevic}, \bibinfo{person}{Charles Ofria}, \bibinfo{person}{Richard~E Lenski}, \bibinfo{person}{Santiago~F Elena}, {and} \bibinfo{person}{Rafael Sanju{\'a}n}.} \bibinfo{year}{2008}\natexlab{}.
\newblock \showarticletitle{Natural selection fails to optimize mutation rates for long-term adaptation on rugged fitness landscapes}.
\newblock \bibinfo{journal}{\emph{PLoS Computational Biology}} \bibinfo{volume}{4}, \bibinfo{number}{9} (\bibinfo{year}{2008}), \bibinfo{pages}{e1000187}.
\newblock


\bibitem[\protect\citeauthoryear{Doerr, Witt, and Yang}{Doerr et~al\mbox{.}}{2018}]%
        {10.1145/3205455.3205569}
\bibfield{author}{\bibinfo{person}{Benjamin Doerr}, \bibinfo{person}{Carsten Witt}, {and} \bibinfo{person}{Jing Yang}.} \bibinfo{year}{2018}\natexlab{}.
\newblock \showarticletitle{Runtime Analysis for Self-Adaptive Mutation Rates}. In \bibinfo{booktitle}{\emph{Proceedings of the Genetic and Evolutionary Computation Conference}} (Kyoto, Japan) \emph{(\bibinfo{series}{GECCO'18})}. \bibinfo{publisher}{Association for Computing Machinery}, \bibinfo{address}{New York, NY, USA}, \bibinfo{pages}{1475–1482}.
\newblock
\showISBNx{9781450356183}
\urldef\tempurl%
\url{https://doi.org/10.1145/3205455.3205569}
\showDOI{\tempurl}


\bibitem[\protect\citeauthoryear{Fialho, Da~Costa, Schoenauer, and Sebag}{Fialho et~al\mbox{.}}{2008}]%
        {10.1007/978-3-540-87700-4_18}
\bibfield{author}{\bibinfo{person}{{\'A}lvaro Fialho}, \bibinfo{person}{Lu{\'i}s Da~Costa}, \bibinfo{person}{Marc Schoenauer}, {and} \bibinfo{person}{Mich{\`e}le Sebag}.} \bibinfo{year}{2008}\natexlab{}.
\newblock \showarticletitle{Extreme Value Based Adaptive Operator Selection}. In \bibinfo{booktitle}{\emph{Parallel Problem Solving from Nature -- PPSN X}}, \bibfield{editor}{\bibinfo{person}{G{\"u}nter Rudolph}, \bibinfo{person}{Thomas Jansen}, \bibinfo{person}{Nicola Beume}, \bibinfo{person}{Simon Lucas}, {and} \bibinfo{person}{Carlo Poloni}} (Eds.). \bibinfo{publisher}{Springer Berlin Heidelberg}, \bibinfo{address}{Berlin, Heidelberg}, \bibinfo{pages}{175--184}.
\newblock
\showISBNx{978-3-540-87700-4}


\bibitem[\protect\citeauthoryear{Glickman and Sycara}{Glickman and Sycara}{2000}]%
        {870276}
\bibfield{author}{\bibinfo{person}{M.R. Glickman} {and} \bibinfo{person}{K. Sycara}.} \bibinfo{year}{2000}\natexlab{}.
\newblock \showarticletitle{Reasons for premature convergence of self-adapting mutation rates}. In \bibinfo{booktitle}{\emph{Proceedings of the 2000 Congress on Evolutionary Computation. CEC00 (Cat. No.00TH8512)}}, Vol.~\bibinfo{volume}{1}. \bibinfo{pages}{62--69}.
\newblock
\urldef\tempurl%
\url{https://doi.org/10.1109/CEC.2000.870276}
\showDOI{\tempurl}


\bibitem[\protect\citeauthoryear{Goldman and Tauritz}{Goldman and Tauritz}{2011}]%
        {10.1145/2001858.2001945}
\bibfield{author}{\bibinfo{person}{Brian~W. Goldman} {and} \bibinfo{person}{Daniel~R. Tauritz}.} \bibinfo{year}{2011}\natexlab{}.
\newblock \showarticletitle{Meta-evolved empirical evidence of the effectiveness of dynamic parameters}. In \bibinfo{booktitle}{\emph{Proceedings of the 13th Annual Conference Companion on Genetic and Evolutionary Computation}} (Dublin, Ireland) \emph{(\bibinfo{series}{GECCO'11})}. \bibinfo{publisher}{Association for Computing Machinery}, \bibinfo{address}{New York, NY, USA}, \bibinfo{pages}{155–156}.
\newblock
\showISBNx{9781450306904}
\urldef\tempurl%
\url{https://doi.org/10.1145/2001858.2001945}
\showDOI{\tempurl}


\bibitem[\protect\citeauthoryear{Helmuth and Kelly}{Helmuth and Kelly}{2021}]%
        {helmuth2021psb2}
\bibfield{author}{\bibinfo{person}{Thomas Helmuth} {and} \bibinfo{person}{Peter Kelly}.} \bibinfo{year}{2021}\natexlab{}.
\newblock \showarticletitle{PSB2: the second program synthesis benchmark suite}. In \bibinfo{booktitle}{\emph{Proceedings of the Genetic and Evolutionary Computation Conference}} (Lille, France) \emph{(\bibinfo{series}{GECCO'21})}. \bibinfo{publisher}{Association for Computing Machinery}, \bibinfo{address}{New York, NY, USA}, \bibinfo{pages}{785–794}.
\newblock
\showISBNx{9781450383509}
\urldef\tempurl%
\url{https://doi.org/10.1145/3449639.3459285}
\showDOI{\tempurl}


\bibitem[\protect\citeauthoryear{Helmuth, McPhee, and Spector}{Helmuth et~al\mbox{.}}{2018}]%
        {helmuth2018program}
\bibfield{author}{\bibinfo{person}{Thomas Helmuth}, \bibinfo{person}{Nicholas~Freitag McPhee}, {and} \bibinfo{person}{Lee Spector}.} \bibinfo{year}{2018}\natexlab{}.
\newblock \showarticletitle{Program synthesis using uniform mutation by addition and deletion}. In \bibinfo{booktitle}{\emph{Proceedings of the Genetic and Evolutionary Computation Conference}} (Kyoto, Japan) \emph{(\bibinfo{series}{GECCO'18})}. \bibinfo{publisher}{Association for Computing Machinery}, \bibinfo{address}{New York, NY, USA}, \bibinfo{pages}{1127–1134}.
\newblock
\showISBNx{9781450356183}
\urldef\tempurl%
\url{https://doi.org/10.1145/3205455.3205603}
\showDOI{\tempurl}


\bibitem[\protect\citeauthoryear{Helmuth, Pantridge, and Spector}{Helmuth et~al\mbox{.}}{2019}]%
        {10.1145/3321707.3321875}
\bibfield{author}{\bibinfo{person}{Thomas Helmuth}, \bibinfo{person}{Edward Pantridge}, {and} \bibinfo{person}{Lee Spector}.} \bibinfo{year}{2019}\natexlab{}.
\newblock \showarticletitle{Lexicase Selection of Specialists}. In \bibinfo{booktitle}{\emph{{\it Proceedings of the Genetic and Evolutionary Computation Conference}}} (Prague, Czech Republic) \emph{(\bibinfo{series}{GECCO'19})}. \bibinfo{publisher}{Association for Computing Machinery}, \bibinfo{address}{New York, NY, USA}, \bibinfo{pages}{1030–1038}.
\newblock
\showISBNx{9781450361118}
\urldef\tempurl%
\url{https://doi.org/10.1145/3321707.3321875}
\showDOI{\tempurl}


\bibitem[\protect\citeauthoryear{Helmuth, Pantridge, and Spector}{Helmuth et~al\mbox{.}}{2020a}]%
        {Helmuth2020}
\bibfield{author}{\bibinfo{person}{Thomas Helmuth}, \bibinfo{person}{Edward Pantridge}, {and} \bibinfo{person}{Lee Spector}.} \bibinfo{year}{2020}\natexlab{a}.
\newblock \showarticletitle{On the importance of specialists for lexicase selection}.
\newblock \bibinfo{journal}{\emph{{\it Genetic Programming and Evolvable Machines}}} \bibinfo{volume}{21}, \bibinfo{number}{3} (\bibinfo{date}{01 Sep} \bibinfo{year}{2020}), \bibinfo{pages}{349--373}.
\newblock
\showISSN{1573-7632}
\urldef\tempurl%
\url{https://doi.org/10.1007/s10710-020-09377-2}
\showDOI{\tempurl}


\bibitem[\protect\citeauthoryear{Helmuth, Pantridge, Woolson, and Spector}{Helmuth et~al\mbox{.}}{2020b}]%
        {10.1162/isal_a_00326}
\bibfield{author}{\bibinfo{person}{Thomas Helmuth}, \bibinfo{person}{Edward Pantridge}, \bibinfo{person}{Grace Woolson}, {and} \bibinfo{person}{Lee Spector}.} \bibinfo{year}{2020}\natexlab{b}.
\newblock \showarticletitle{Genetic Source Sensitivity and Transfer Learning in Genetic Programming}. In \bibinfo{booktitle}{\emph{Proceedings of ALIFE 2020: The 2020 Conference on Artificial Life}}. \bibinfo{pages}{303--311}.
\newblock
\urldef\tempurl%
\url{https://doi.org/10.1162/isal_a_00326}
\showDOI{\tempurl}


\bibitem[\protect\citeauthoryear{Helmuth and Spector}{Helmuth and Spector}{2015}]%
        {helmuth2015general}
\bibfield{author}{\bibinfo{person}{Thomas Helmuth} {and} \bibinfo{person}{Lee Spector}.} \bibinfo{year}{2015}\natexlab{}.
\newblock \showarticletitle{General Program Synthesis Benchmark Suite}. In \bibinfo{booktitle}{\emph{Proceedings of the 2015 Annual Conference on Genetic and Evolutionary Computation}} (Madrid, Spain) \emph{(\bibinfo{series}{GECCO'15})}. \bibinfo{publisher}{Association for Computing Machinery}, \bibinfo{address}{New York, NY, USA}, \bibinfo{pages}{1039–1046}.
\newblock
\showISBNx{9781450334723}
\urldef\tempurl%
\url{https://doi.org/10.1145/2739480.2754769}
\showDOI{\tempurl}


\bibitem[\protect\citeauthoryear{Helmuth, Spector, and Matheson}{Helmuth et~al\mbox{.}}{2015}]%
        {6920034}
\bibfield{author}{\bibinfo{person}{Thomas Helmuth}, \bibinfo{person}{Lee Spector}, {and} \bibinfo{person}{James Matheson}.} \bibinfo{year}{2015}\natexlab{}.
\newblock \showarticletitle{Solving Uncompromising Problems With Lexicase Selection}.
\newblock \bibinfo{journal}{\emph{IEEE Transactions on Evolutionary Computation}} \bibinfo{volume}{19}, \bibinfo{number}{5} (\bibinfo{year}{2015}), \bibinfo{pages}{630--643}.
\newblock
\urldef\tempurl%
\url{https://doi.org/10.1109/TEVC.2014.2362729}
\showDOI{\tempurl}


\bibitem[\protect\citeauthoryear{Kruisselbrink, Li, Reehuis, Eggermont, and B\"{a}ck}{Kruisselbrink et~al\mbox{.}}{2011}]%
        {10.1145/2001576.2001699}
\bibfield{author}{\bibinfo{person}{Johannes~W. Kruisselbrink}, \bibinfo{person}{Rui Li}, \bibinfo{person}{Edgar Reehuis}, \bibinfo{person}{Jeroen Eggermont}, {and} \bibinfo{person}{Thomas B\"{a}ck}.} \bibinfo{year}{2011}\natexlab{}.
\newblock \showarticletitle{On the log-normal self-adaptation of the mutation rate in binary search spaces}. In \bibinfo{booktitle}{\emph{Proceedings of the 13th Annual Conference on Genetic and Evolutionary Computation}} (Dublin, Ireland) \emph{(\bibinfo{series}{GECCO'11})}. \bibinfo{publisher}{Association for Computing Machinery}, \bibinfo{address}{New York, NY, USA}, \bibinfo{pages}{893–900}.
\newblock
\showISBNx{9781450305570}
\urldef\tempurl%
\url{https://doi.org/10.1145/2001576.2001699}
\showDOI{\tempurl}


\bibitem[\protect\citeauthoryear{Kuleshov and Precup}{Kuleshov and Precup}{2014}]%
        {kuleshov2014algorithms}
\bibfield{author}{\bibinfo{person}{Volodymyr Kuleshov} {and} \bibinfo{person}{Doina Precup}.} \bibinfo{year}{2014}\natexlab{}.
\newblock \bibinfo{title}{Algorithms for multi-armed bandit problems}.
\newblock
\newblock
\showeprint[arxiv]{1402.6028}~[cs.AI]


\bibitem[\protect\citeauthoryear{Kumar, Liu, Miikkulainen, and Stone}{Kumar et~al\mbox{.}}{2022}]%
        {kumar2022effective}
\bibfield{author}{\bibinfo{person}{Akarsh Kumar}, \bibinfo{person}{Bo Liu}, \bibinfo{person}{Risto Miikkulainen}, {and} \bibinfo{person}{Peter Stone}.} \bibinfo{year}{2022}\natexlab{}.
\newblock \showarticletitle{Effective mutation rate adaptation through group elite selection}. In \bibinfo{booktitle}{\emph{Proceedings of the Genetic and Evolutionary Computation Conference}} (Boston, Massachusetts) \emph{(\bibinfo{series}{GECCO'22})}. \bibinfo{publisher}{Association for Computing Machinery}, \bibinfo{address}{New York, NY, USA}, \bibinfo{pages}{721–729}.
\newblock
\showISBNx{9781450392372}
\urldef\tempurl%
\url{https://doi.org/10.1145/3512290.3528706}
\showDOI{\tempurl}


\bibitem[\protect\citeauthoryear{La~Cava, Spector, and Danai}{La~Cava et~al\mbox{.}}{2016}]%
        {10.1145/2908812.2908898}
\bibfield{author}{\bibinfo{person}{William La~Cava}, \bibinfo{person}{Lee Spector}, {and} \bibinfo{person}{Kourosh Danai}.} \bibinfo{year}{2016}\natexlab{}.
\newblock \showarticletitle{Epsilon-Lexicase Selection for Regression}. In \bibinfo{booktitle}{\emph{Proceedings of the Genetic and Evolutionary Computation Conference 2016}} (Denver, Colorado, USA) \emph{(\bibinfo{series}{GECCO'16})}. \bibinfo{publisher}{Association for Computing Machinery}, \bibinfo{address}{New York, NY, USA}, \bibinfo{pages}{741–748}.
\newblock
\showISBNx{9781450342063}
\urldef\tempurl%
\url{https://doi.org/10.1145/2908812.2908898}
\showDOI{\tempurl}


\bibitem[\protect\citeauthoryear{Malan and Engelbrecht}{Malan and Engelbrecht}{2009}]%
        {4983112}
\bibfield{author}{\bibinfo{person}{Katherine~M. Malan} {and} \bibinfo{person}{Andries~P. Engelbrecht}.} \bibinfo{year}{2009}\natexlab{}.
\newblock \showarticletitle{Quantifying ruggedness of continuous landscapes using entropy}. In \bibinfo{booktitle}{\emph{2009 IEEE Congress on Evolutionary Computation}}. \bibinfo{pages}{1440--1447}.
\newblock
\urldef\tempurl%
\url{https://doi.org/10.1109/CEC.2009.4983112}
\showDOI{\tempurl}


\bibitem[\protect\citeauthoryear{Marsili~Libelli and Alba}{Marsili~Libelli and Alba}{2000}]%
        {marsili2000adaptive}
\bibfield{author}{\bibinfo{person}{Stefano Marsili~Libelli} {and} \bibinfo{person}{P Alba}.} \bibinfo{year}{2000}\natexlab{}.
\newblock \showarticletitle{Adaptive mutation in genetic algorithms}.
\newblock \bibinfo{journal}{\emph{Soft computing}}  \bibinfo{volume}{4} (\bibinfo{year}{2000}), \bibinfo{pages}{76--80}.
\newblock


\bibitem[\protect\citeauthoryear{Maschek}{Maschek}{2010}]%
        {Maschek2010}
\bibfield{author}{\bibinfo{person}{Michael~Kurtis Maschek}.} \bibinfo{year}{2010}\natexlab{}.
\newblock \showarticletitle{Intelligent Mutation Rate Control in an Economic Application of Genetic Algorithms}.
\newblock \bibinfo{journal}{\emph{Computational Economics}} \bibinfo{volume}{35}, \bibinfo{number}{1} (\bibinfo{date}{01 Jan} \bibinfo{year}{2010}), \bibinfo{pages}{25--49}.
\newblock
\showISSN{1572-9974}
\urldef\tempurl%
\url{https://doi.org/10.1007/s10614-009-9190-6}
\showDOI{\tempurl}


\bibitem[\protect\citeauthoryear{Metzgar and Wills}{Metzgar and Wills}{2000}]%
        {metzgar2000evidence}
\bibfield{author}{\bibinfo{person}{David Metzgar} {and} \bibinfo{person}{Christopher Wills}.} \bibinfo{year}{2000}\natexlab{}.
\newblock \showarticletitle{Evidence for the adaptive evolution of mutation rates}.
\newblock \bibinfo{journal}{\emph{Cell}} \bibinfo{volume}{101}, \bibinfo{number}{6} (\bibinfo{year}{2000}), \bibinfo{pages}{581--584}.
\newblock


\bibitem[\protect\citeauthoryear{Meyer-Nieberg and Beyer}{Meyer-Nieberg and Beyer}{2007}]%
        {meyer2007self}
\bibfield{author}{\bibinfo{person}{Silja Meyer-Nieberg} {and} \bibinfo{person}{Hans-Georg Beyer}.} \bibinfo{year}{2007}\natexlab{}.
\newblock \showarticletitle{Self-adaptation in evolutionary algorithms}.
\newblock In \bibinfo{booktitle}{\emph{Parameter setting in evolutionary algorithms}}. \bibinfo{publisher}{Springer}, \bibinfo{pages}{47--75}.
\newblock


\bibitem[\protect\citeauthoryear{Poli, Langdon, and McPhee}{Poli et~al\mbox{.}}{2008}]%
        {fieldguidegp}
\bibfield{author}{\bibinfo{person}{Ricardo Poli}, \bibinfo{person}{William~B. Langdon}, {and} \bibinfo{person}{Nicholas~F. McPhee}.} \bibinfo{year}{2008}\natexlab{}.
\newblock \bibinfo{booktitle}{\emph{A Field Guide to Genetic Programming}}.
\newblock \bibinfo{publisher}{Lulu Enterprises}.
\newblock


\bibitem[\protect\citeauthoryear{Spector, Perry, Klein, and Keijzer}{Spector et~al\mbox{.}}{2004}]%
        {spector2004push}
\bibfield{author}{\bibinfo{person}{Lee Spector}, \bibinfo{person}{Chris Perry}, \bibinfo{person}{Jon Klein}, {and} \bibinfo{person}{Maarten Keijzer}.} \bibinfo{year}{2004}\natexlab{}.
\newblock \bibinfo{booktitle}{\emph{Push 3.0 programming language description}}.
\newblock
\urldef\tempurl%
\url{https://faculty.hampshire.edu/lspector/temp/HC-CSTR-2004-02.pdf}
\showURL{%
Retrieved Jan 30, 2024 from \tempurl}


\bibitem[\protect\citeauthoryear{Spector and Robinson}{Spector and Robinson}{2002}]%
        {Spector2002}
\bibfield{author}{\bibinfo{person}{Lee Spector} {and} \bibinfo{person}{Alan Robinson}.} \bibinfo{year}{2002}\natexlab{}.
\newblock \showarticletitle{Genetic Programming and Autoconstructive Evolution with the Push Programming Language}.
\newblock \bibinfo{journal}{\emph{Genetic Programming and Evolvable Machines}} \bibinfo{volume}{3}, \bibinfo{number}{1} (\bibinfo{date}{01 Mar} \bibinfo{year}{2002}), \bibinfo{pages}{7--40}.
\newblock
\showISSN{1573-7632}
\urldef\tempurl%
\url{https://doi.org/10.1023/A:1014538503543}
\showDOI{\tempurl}


\bibitem[\protect\citeauthoryear{Stephens, Olmedo, Vargas, and Waelbroeck}{Stephens et~al\mbox{.}}{1998}]%
        {6788145}
\bibfield{author}{\bibinfo{person}{C.~R. Stephens}, \bibinfo{person}{I.~García Olmedo}, \bibinfo{person}{J.~Mora Vargas}, {and} \bibinfo{person}{H. Waelbroeck}.} \bibinfo{year}{1998}\natexlab{}.
\newblock \showarticletitle{Self-Adaptation in Evolving Systems}.
\newblock \bibinfo{journal}{\emph{Artificial Life}} \bibinfo{volume}{4}, \bibinfo{number}{2} (\bibinfo{year}{1998}), \bibinfo{pages}{183--201}.
\newblock
\urldef\tempurl%
\url{https://doi.org/10.1162/106454698568512}
\showDOI{\tempurl}


\bibitem[\protect\citeauthoryear{Surjanovic and Bingham}{Surjanovic and Bingham}{2013}]%
        {simlib}
\bibfield{author}{\bibinfo{person}{Sonja Surjanovic} {and} \bibinfo{person}{Derek Bingham}.} \bibinfo{year}{2013}\natexlab{}.
\newblock \bibinfo{booktitle}{\emph{Optimization Test Functions and Datasets}}.
\newblock
\urldef\tempurl%
\url{https://www.sfu.ca/~ssurjano/optimization.html}
\showURL{%
Retrieved Jan 28, 2024 from \tempurl}


\bibitem[\protect\citeauthoryear{Sutskever, Martens, Dahl, and Hinton}{Sutskever et~al\mbox{.}}{2013}]%
        {pmlr-v28-sutskever13}
\bibfield{author}{\bibinfo{person}{Ilya Sutskever}, \bibinfo{person}{James Martens}, \bibinfo{person}{George Dahl}, {and} \bibinfo{person}{Geoffrey Hinton}.} \bibinfo{year}{2013}\natexlab{}.
\newblock \showarticletitle{On the importance of initialization and momentum in deep learning}. In \bibinfo{booktitle}{\emph{Proceedings of the 30th International Conference on Machine Learning}} \emph{(\bibinfo{series}{Proceedings of Machine Learning Research}, Vol.~\bibinfo{volume}{28})}, \bibfield{editor}{\bibinfo{person}{Sanjoy Dasgupta} {and} \bibinfo{person}{David McAllester}} (Eds.). \bibinfo{publisher}{PMLR}, \bibinfo{address}{Atlanta, Georgia, USA}, \bibinfo{pages}{1139--1147}.
\newblock
\urldef\tempurl%
\url{https://proceedings.mlr.press/v28/sutskever13.html}
\showURL{%
\tempurl}


\bibitem[\protect\citeauthoryear{Sutton and Barto}{Sutton and Barto}{2018}]%
        {sutton2018reinforcement}
\bibfield{author}{\bibinfo{person}{Richard~S Sutton} {and} \bibinfo{person}{Andrew~G Barto}.} \bibinfo{year}{2018}\natexlab{}.
\newblock \bibinfo{booktitle}{\emph{Reinforcement learning: An introduction}}.
\newblock \bibinfo{publisher}{MIT press}.
\newblock


\bibitem[\protect\citeauthoryear{Uy, Hoai, O'Neill, McKay, and Galv{\'a}n-L{\'o}pez}{Uy et~al\mbox{.}}{2011}]%
        {Uy2011}
\bibfield{author}{\bibinfo{person}{Nguyen~Quang Uy}, \bibinfo{person}{Nguyen~Xuan Hoai}, \bibinfo{person}{Michael O'Neill}, \bibinfo{person}{R.~I. McKay}, {and} \bibinfo{person}{Edgar Galv{\'a}n-L{\'o}pez}.} \bibinfo{year}{2011}\natexlab{}.
\newblock \showarticletitle{Semantically-based crossover in genetic programming: application to real-valued symbolic regression}.
\newblock \bibinfo{journal}{\emph{Genetic Programming and Evolvable Machines}} \bibinfo{volume}{12}, \bibinfo{number}{2} (\bibinfo{date}{01 Jun} \bibinfo{year}{2011}), \bibinfo{pages}{91--119}.
\newblock
\showISSN{1573-7632}
\urldef\tempurl%
\url{https://doi.org/10.1007/s10710-010-9121-2}
\showDOI{\tempurl}


\bibitem[\protect\citeauthoryear{Whitacre, Pham, and Sarker}{Whitacre et~al\mbox{.}}{2006}]%
        {10.1145/1143997.1144205}
\bibfield{author}{\bibinfo{person}{James~M. Whitacre}, \bibinfo{person}{Tuan~Q. Pham}, {and} \bibinfo{person}{Ruhul~A. Sarker}.} \bibinfo{year}{2006}\natexlab{}.
\newblock \showarticletitle{Use of statistical outlier detection method in adaptive evolutionary algorithms}. In \bibinfo{booktitle}{\emph{Proceedings of the 8th Annual Conference on Genetic and Evolutionary Computation}} (Seattle, Washington, USA) \emph{(\bibinfo{series}{GECCO'06})}. \bibinfo{publisher}{Association for Computing Machinery}, \bibinfo{address}{New York, NY, USA}, \bibinfo{pages}{1345–1352}.
\newblock
\showISBNx{1595931864}
\urldef\tempurl%
\url{https://doi.org/10.1145/1143997.1144205}
\showDOI{\tempurl}


\bibitem[\protect\citeauthoryear{White, McDermott, Castelli, Manzoni, Goldman, Kronberger, Ja{\'{s}}kowski, O'Reilly, and Luke}{White et~al\mbox{.}}{2013}]%
        {betterGPBenchmarks}
\bibfield{author}{\bibinfo{person}{David~R. White}, \bibinfo{person}{James McDermott}, \bibinfo{person}{Mauro Castelli}, \bibinfo{person}{Luca Manzoni}, \bibinfo{person}{Brian~W. Goldman}, \bibinfo{person}{Gabriel Kronberger}, \bibinfo{person}{Wojciech Ja{\'{s}}kowski}, \bibinfo{person}{Una-May O'Reilly}, {and} \bibinfo{person}{Sean Luke}.} \bibinfo{year}{2013}\natexlab{}.
\newblock \showarticletitle{Better GP benchmarks: community survey results and proposals}.
\newblock \bibinfo{journal}{\emph{Genetic Programming and Evolvable Machines}} \bibinfo{volume}{14}, \bibinfo{number}{1} (\bibinfo{date}{01 Mar} \bibinfo{year}{2013}), \bibinfo{pages}{3--29}.
\newblock
\showISSN{1573-7632}
\urldef\tempurl%
\url{https://doi.org/10.1007/s10710-012-9177-2}
\showDOI{\tempurl}


\end{thebibliography}

\appendix

\section{Test Function Definitions}

In this section we describe the test functions used in our function minimization experiment as well as the standard deviation used to initialize our population. Aside from the Linear function, which is unbounded, and the Rosenbrock function, which achieves its minimum at ${\bf x}={\bf 1}$, all test functions are constructed to have a global minimum of 0 at ${\bf x}={\bf 0}$. In these definitions, $d$ is the dimension of the vector $\bf{x}$ and the $x_i$ are all 1-indexed.

\subsection{Ackley}

\begin{equation}
    f({\bf x})=-a\exp\left(-b\sqrt{\frac{1}{d}\sum_{i=1}^dx_i^2}\right)-\exp\left(\frac{1}{d}\sum_{i=1}^d\cos(cx_i)\right)+a+e
\end{equation}

where $a=20$, $b=0.2$, and $c=2\pi$.

As Surjanovic et al. \cite{simlib} recommend a search range of $x_i\in [-32.768, 32.768]$ for $1\le i\le d$, we initialize the population with a standard deviation of 10 for this problem.

\subsection{Griewank}

\begin{equation}
    f({\bf x})=\sum_{i=1}^d\frac{x_i^2}{4000}-\prod_{i=1}^d\cos\left(\frac{x_i}{\sqrt{i}}\right)+1
\end{equation}

As Surjanovic et al. \cite{simlib} recommend a search range of $x_i\in [-600, 600]$ for $1\le i\le d$, we initialize the population with a standard deviation of 1000 for this problem.

\subsection{Rastrigin}

\begin{equation}
    f({\bf x})=10d+\sum_{i=1}^d[x_i^2-10\cos(2\pi x_i)]
\end{equation}

As Surjanovic et al. \cite{simlib} recommend a search range of $x_i\in [-5.12, 5.12]$ for $1\le i\le d$, we initialize the population with a standard deviation of 10 for this problem.

\subsection{Rosenbrock}

\begin{equation}
    f({\bf x})=\sum_{i=1}^{d-1}[100(x_{i+1}-x_i^2)^2+(x_i-1)^2]
\end{equation}

As Surjanovic et al. \cite{simlib} recommend a search range of $x_i\in [-2.048, 2.048]$ for $1\le i\le d$, we initialize the population with a standard deviation of 1 for this problem.

\subsection{Sphere}

\begin{equation}
    f({\bf x})=\sum_{i=1}^dx_i^2
\end{equation}

As Surjanovic et al. \cite{simlib} recommend a search range of $x_i\in [-5.12, 5.12]$ for $1\le i\le d$, we initialize the population with a standard deviation of 10 for this problem.

\subsection{Linear}

\begin{equation}
    f({\bf x})=\sum_{i=1}^dx_i
\end{equation}

This problem is not detailed in Surjanovic et al. \cite{simlib}, so we use a default standard deviation of 1.

\section{Symbolic Regression Function Definitions}
In our Symbolic Regression experiments, we attempt to recover a known mathematical function $f(x)$ using a genetic algorithm based on some example input-output pairs $(x_i, f(x_i))$. 
In this section we detail the specific target functions and input ranges used, following the format of White et al.\cite{betterGPBenchmarks}.
\subsection{Nguyen1}

\begin{equation}
    f(x)=x^3+x^2+x
\end{equation}
For this problem our input is a range of evenly spaced floats from $-4$ to $4$ with step size $0.1$: $E[-4.0,4.0,0.1]$
\subsection{Nguyen2}
\begin{equation}
    f(x)=x^4+x^3+x^2+x
\end{equation}
For this problem our input is $E[-4.0,4.0,0.1]$
\subsection{Nguyen3}
\begin{equation}
    f(x)=x^5+x^4+x^3+x^2+x
\end{equation}
For this problem our input is $E[-4.0,4.0,0.1]$
\subsection{Nguyen4}
\begin{equation}
    f(x)=x^6+x^5+x^4+x^3+x^2+x
\end{equation}
For this problem our input is $E[-4.0,4.0,0.1]$
\subsection{Nguyen5}
\begin{equation}
    f(x)=\sin(x^2)\cos(x)-1
\end{equation}
For this problem our input is $E[-4.0,4.0,0.1]$
\subsection{Nguyen6}
\begin{equation}
    f(x)=\sin(x)+\sin(x+x^2)
\end{equation}
For this problem our input is $E[-4.0,4.0,0.1]$
\subsection{Nguyen7}
\begin{equation}
    f(x)=\log(x+1)+\log(x^2+1)
\end{equation}
As this function is undefined for inputs smaller than $-1$, we use $E[0.0,8.0,0.1]$ as inputs.
\subsection{Nguyen8}
\begin{equation}
    f(x)=\sqrt{x}
\end{equation}
As this function is undefined for negative inputs, we use $E[0.0,8.0,0.1]$ as inputs.

\section{Parameter Settings}

The parameter settings used by our adaptive controller are shown in Table \ref{tab:parameters}. We use an epsilon-greedy strategy with sampling noise for exploration. Each bandit has a randomized learning rate. For each bandit, we create 20 tile codings, each with tile width and tile offset sampled uniformly from the range of possible values. In order to be able to evaluate each tile in the base coding, we require the tile widths and offsets to all be multiples of the base coding tile width.

We use a value of 7 for our sampling noise in the function minimization problems. The corresponding noise parameter in GESMR is the meta-mutation strength, which they set to 2. In GESMR, mutation rates are mutated by multiplying by $2^{\mathcal{U}[-1,1]}$. Since the random variable $\log(2^{\mathcal{U}([-1,1])})$ has standard deviation $\sqrt{\frac{2\log(2)}{12}}=0.223$, we set the standard deviation of our sampled UMAD rates to $0.223$ as well. Since our base tile width is 0.03, this comes out to a standard deviation in terms of tile indices of $\frac{0.223}{0.03}\approx7$

While we also have to specify the added constant $c$ in the proxy error function for each problem, we set this to around the minimum absolute nonzero error. For software synthesis, errors are integer-valued, so we set the constant to 1. For symbolic regression, Nguyen et al. \cite{Uy2011} considers any error below 0.01 a ``hit,'' so we set our constant to 0.01. Finally, since function minimization errors are 1-dimensional, we do not need to use this lexicase-like error transformation step.

\begin{table*}[]
\caption{Hyperparameter values for bandit controller}
    \label{tab:parameters}
    \begin{tabular}{|c|c|}\toprule
        Parameter & Value \\\midrule
        Number of Bandits in Ensemble & 5\\
        $len\_history$ & 100\\ 
        Learning Rate $\gamma$ & $\gamma\sim10^{\mathcal{U}([-4,-3])}$\\ 
        Momentum Factor $\mu$ & 0.9 \\
        Epsilon $\epsilon$ & Annealed linearly from 1 to 0.01 over 5 generations\\
        \multirow{2}{*}{Sampling noise $\sigma$}  & 3 (software synthesis and SR) \\ 
        & 7 (function minimization)\\
          Number of tile codings & 20\\
        Log-UMAD interval  & [-10, 0] (software synthesis and SR)\\ Log-$\sigma$ interval& [-100, 100] (function minimization)\\
        Base coding tile width & 0.03 \\
        Tile coding widths & $w_i\sim\text{Uniform}(\{0.18, 0.21, 0.24, 0.27, 0.3, 0.33, 0.36, 0.39\})$\\
        Tile coding offsets & $o_i\sim\text{Uniform}(\{0, 0.03, 0.06, 0.09, 0.12, 0.15\})$\\
      
        \multirow{3}{*}{Proxy error function}
        & $f(x)=\text{sgn}(x)\log(1+|x|)$ (software synthesis)\\
        & $f(x)=\text{sgn}(x)\log(0.01+|x|)$ (symbolic regression)\\
        & $f(x)=x$ (function minimization)\\\bottomrule
    \end{tabular}
\end{table*}

The hyperparameters used in the function minimization experiments are detailed in table \ref{tab:parameters}. Following the recommendation in \cite{kumar2022effective}, we set the GESMR meta population size to be the largest divisor of $N-1$ (the number of mutated individuals) that is less than $N^{\frac{3}{4}}$. Aside from the GA truncation size and the GESMR meta truncation size, which were not available in \cite{kumar2022effective} and are therefore set to reasonable values, we take all other hyperparameters from Kumar et al.\cite{kumar2022effective}
\begin{table*}[]
\caption{Hyperparameter values for Function Minimization}
    \label{tab:parameters}
    \begin{tabular}{|c|c|}\toprule
        Parameter & Value \\\midrule
        Population size $N$ & 100 + 1 elite\\
        Selection operator & Truncation selection\\
        Truncation size & 10 \\
        GESMR meta selection operator & Truncation selection \\
        GESMR meta truncation size & 4\\
        GESMR meta population size & 9 + 1 elite \\
        GESMR meta mutation rate & 2 \\
        SAMR meta mutation rate & 2 \\
        GESMR rate initialization & 10 Log-spaced values from $10^{-3}$ to $10^3$ \\
        SAMR rate initialization & 101 Log-spaced from $10^{-3}$ to $10^3$ \\
        LAMR searched rates & 10 log-spaced values from $10^{-3}$ to $10^0$\\
        LAMR lookahead generation & 100 
        \\\bottomrule
    \end{tabular}
    
\end{table*}

\end{document}